\documentclass[letterpaper, 10 pt, conference]{ieeeconf}  

\IEEEoverridecommandlockouts                              

\usepackage{amsmath}
\usepackage{cite}
\usepackage{graphicx}
\usepackage{subcaption}
\usepackage{tikz}
\usetikzlibrary{arrows.meta, positioning, shapes, calc}
\usepackage{amssymb}
\usepackage{multirow}

\title{\LARGE \bf
Robust Visual Embodiment: How Robots Discover Their Bodies in Real Environments}

\author{Salim Rezvani, Ammar Jaleel Mahmood, and Robin Chhabra
\thanks{*This work was not supported by any organization}
\thanks{$^{1}$Robin Chhabra is with the Department of Mechanical, Industrial and Mechatronics Engineering, Toronto Metropolitan University, Toronto, Canada
        {\tt\small robin.chhabra@torontomu.ca}}%
\thanks{
        {\tt\small salim.rezvani@torontomu.ca}}%
\thanks{
        {\tt\small ammar.j.mahmood@torontomu.ca}}%
}

\begin{document}

\maketitle
\thispagestyle{empty}
\pagestyle{empty}


\begin{abstract}

Robots with internal visual self-models promise unprecedented adaptability, yet existing autonomous modeling pipelines remain fragile under realistic sensing conditions such as noisy imagery and cluttered backgrounds. This paper presents the first systematic study quantifying how visual degradations—including blur, salt-and-pepper noise, and Gaussian noise—affect robotic self-modeling. Through both simulation and physical experiments, we demonstrate their impact on morphology prediction, trajectory planning, and damage recovery in state-of-the-art pipelines. To overcome these challenges, we introduce a task-aware denoising framework that couples classical restoration with morphology-preserving constraints, ensuring retention of structural cues critical for self-modeling. In addition, we integrate semantic segmentation to robustly isolate robots from cluttered and colorful scenes. Extensive experiments show that our approach restores near-baseline performance across simulated and physical platforms, while existing pipelines degrade significantly. These contributions advance the robustness of visual self-modeling and establish practical foundations for deploying self-aware robots in unpredictable real-world environments.

\end{abstract}

\section{Introduction}

Self-supervised robotic self-modeling enables machines to autonomously infer their morphology and kinematics directly from visual data, without relying on pre-defined models \cite{1,2,3}. By observing themselves through onboard or external cameras, robots can iteratively refine predictive models of their bodies and use these models for motion planning, adaptation, and, critically, damage recovery during long-duration missions in remote and harsh environments such as outer space. However, despite recent advances, most existing approaches rely on highly idealized conditions including noise-free inputs, static monochrome backgrounds, and controlled laboratory settings that significantly constrain their applicability in real-world environments.

In practice, robotic vision systems are inevitably exposed to degradations. Motion blur arises from camera or robot movement, salt-and-pepper noise from sensor faults or dust, and Gaussian noise from electronic interference \cite{6,7,8,9,10}. Such perturbations propagate through self-modeling pipelines, distorting morphology inference and reducing downstream task performance \cite{11,12,13,14}. While the robotics community has extensively studied actuator and sensor noise \cite{4,5}, the impact of visual noise on robotic self-modeling remains largely unexplored. This gap is critical since self-modeling largely depends on high-quality visual input. 

Classical denoising methods including median filtering, Wiener filtering, and Non-Local Means \cite{15,16,17,18,19,20,21,22,23,24,25,26,27,28} were developed for generic image restoration. However, robotic self-modeling requires not only clean images, but faithful preservation of kinematic chains and joint boundaries. Standard filters may remove noise yet erase precisely the structural cues needed for prediction. For example, a median filter may suppress salt-and-pepper noise but distort limb contours, while Gaussian blur can obscure fine geometry essential for motion prediction. These limitations motivate task-aware denoising strategies specifically tailored for self-modeling.

The Intuitionistic Fuzzy Twin Support Vector Machine (IFT-SVM), introduced in \cite{36,37,38}, extends Twin Support Vector Machines by incorporating intuitionistic fuzzy sets, assigning both membership and non-membership degrees to data points. This dual representation allows IFT-SVM to robustly handle ambiguous or noisy sensor (image) data by distinguishing reliable information from outliers. In our pipeline, IFT-SVM is integrated with Non-Local Means (NLM) denoising to refine pixel neighborhoods adaptively, reducing residual Gaussian noise while preserving fine structures. By directly improving the quality of visual input, IFT-SVM ensures that the downstream robotic self-modeling stage receives cleaner, more reliable morphology cues. Its ability to manage uncertainty thus makes it a natural fit for noise-resilient self-modeling, where maintaining high-quality visual data is critical for accurate reconstruction and decision-making.

The Free-Form Kinematic Self-Model (FFKSM) proposed by Hu \textit{et al.} \cite{1} demonstrated that robots can predict morphology and kinematics without predefined equations, CAD models, or multi-sensor setups. FFKSM employs a coordinate encoder, kinematic encoder, and predictive module trained via self-supervision on binary segmentation masks. Subsequent work extended this direction: Xie \textit{et al.} \cite{34} introduced a voxel-based framework capturing dense 3D geometry, while Back \textit{et al.} \cite{35} combined semantic segmentation with predictive dynamics to model novel appendages or tools. While these systems improved geometric fidelity and flexibility, they largely assumed clean inputs and controlled backgrounds. Noise-robust, segmentation-aware self-modeling remains an open challenge.

Another important limitation in the current literature is the assumption of uniform backgrounds. For example, FFKSM \cite{1} relies on color-based segmentation masks to bootstrap training. While effective in controlled labs, this assumption breaks in cluttered, colorful, or dynamic scenes. Early color-based methods \cite{29,30} are similarly fragile. Recent advances in semantic segmentation address these challenges: Xie \textit{et al.} \cite{31} introduced RICE for refining instance masks in cluttered scenes, Back \textit{et al.} \cite{32} developed boundary refinement for unknown objects, and Wu \textit{et al.} \cite{33} applied transformer-based segmentation for robotics. These approaches achieve robust extraction in cluttered, occluded, or dynamic environments, suggesting a path beyond the fragile white-background assumption.

To address these challenges, we introduce a task-aware denoising pipeline that restores corrupted visual inputs while maintaining morphological fidelity. By integrating noise suppression with structured self-modeling, we systematically quantify the effects of three representative noise types—blur, salt-and-pepper, and Gaussian—on robotic systems. Unlike perception-focused approaches \cite{5,8,10}, which primarily aim at recognizing objects or scenes under noisy conditions, our framework is designed specifically for self-modeling, where preserving accurate robot morphology is critical. Furthermore, by combining this denoising pipeline with semantic segmentation, the two most pervasive real-world challenges, our method extends beyond controlled setups and enables robust self-modeling in cluttered, realistic environments, outside the laboratory. 
Our contributions are threefold:

1) \textbf{Noise-Aware Benchmarking}: We present the first systematic robustness evaluation of self-modeling under various visual noise types on both simulated and physical robots.  

2) \textbf{Task-Aware Denoising}: We design a modular denoising framework combining Wiener filtering, median filtering, and Non-Local Means, explicitly preserving morphological fidelity rather than focusing only on pixel-level accuracy.  

3) \textbf{Semantic Segmentation for Real-World Deployment}: We incorporate semantic segmentation into the self-modeling pipeline, enabling robust robot isolation in cluttered and colorful backgrounds. 

Our experiments on simulated robots and 3D-printed prototypes show that state-of-the-art self-modeling pipelines such as FFKSM degrade sharply under noise and clutter, whereas our enhanced framework restores near-baseline accuracy.

The rest of the paper is organized as follows: Section \ref{sec:noise_simulation} details the noise types considered in this work. We propose our self-modeling technique in Section \ref{sec:denoising_techniques}. Our experimental setup is explained in Section \ref{exp-setup} with results reported in Section \ref{exp-results}. Section \ref{concl} includes some concluding remarks.

\section{Noise Types}
\label{sec:noise_simulation}


Images captured in robotic perception are often affected by noise due to environmental factors, sensor imperfections, or motion. To study robustness under such conditions, we consider three representative types of noise in our experiments: blur, salt-and-pepper noise, and Gaussian noise.

\subsection{Blur}

Blurring reduces edge sharpness and fine structures, often caused by camera motion, lens imperfections, or environmental factors. We model blur using a Gaussian kernel:

\begin{equation}
I_\text{blur}(x, y) = (I * G_\sigma)(x, y) = \sum_{i=-k}^{k} \sum_{j=-k}^{k} I(x-i, y-j) G_\sigma(i,j),
\end{equation}
with
\begin{equation}
G_\sigma(i,j) = \frac{1}{2 \pi \sigma^2} \exp\left(-\frac{i^2+j^2}{2\sigma^2}\right),
\end{equation}
where $k$ determines the kernel size and $\sigma$ controls the extent of blurring. This degradation reduces edge visibility and fine structures critical for self-modeling.

\subsection{Salt-and-Pepper Noise}
Salt-and-pepper noise introduces random extreme pixel values, either black (0) or white (255), across an image:

\begin{equation}
I_\text{sp}(x, y) =
\begin{cases}
0 & \text{with probability } p/2,\\
255 & \text{with probability } p/2,\\
I(x, y) & \text{with probability } 1-p,
\end{cases}
\end{equation}
where $p$ is the noise density. This impulsive noise disrupts color-based segmentation and adds high-frequency disturbances that challenge visual self-modeling.

\subsection{Gaussian (Artificial) Noise}

Gaussian noise simulates common real-world sensor perturbations, such as electronic interference or low-light sensor variability. We model it as additive zero-mean Gaussian:

\begin{equation}
I_\text{gauss}(x, y) = \text{clip}\Big(I(x, y) + N_G(x,y), 0, 255 \Big),
\end{equation}
where $N_G(x,y) \sim \mathcal{N}(0, \sigma^2)$ and $\text{clip}(\cdot)$ ensures pixel values remain within the valid range. This allows assessment of self-modeling robustness under typical sensor noise conditions.



\section{Proposed Self-Modeling Pipeline}
\label{sec:denoising_techniques}

\begin{figure*}[!t]
\centering
\includegraphics[width=6in, height=2.5in]{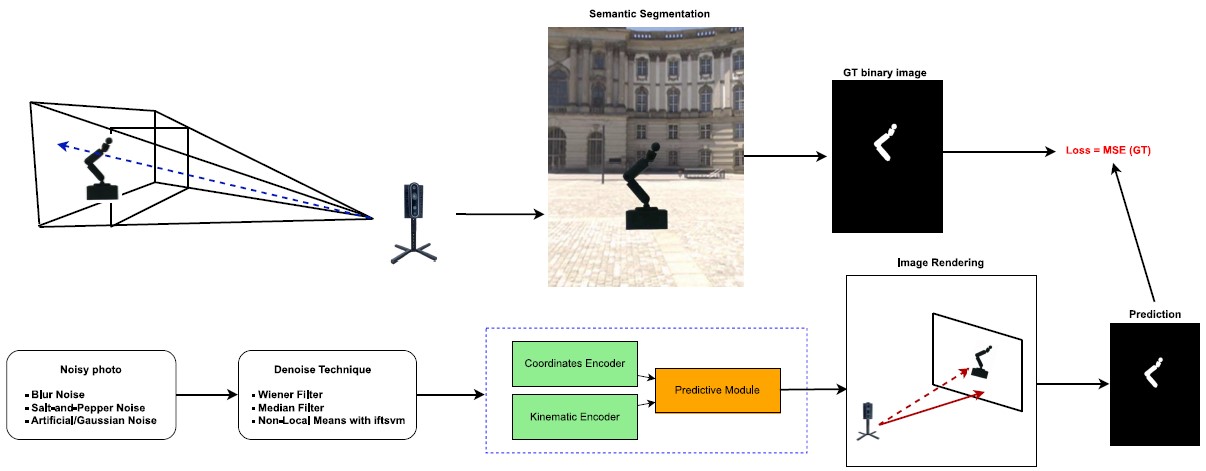}
\caption{Self-modeling pipeline overview}
\label{Pipeline}
\end{figure*}

While noise primarily perturbs pixel-level information, real-world scenes introduce additional challenges, such as cluttered or colorful backgrounds that can obscure robot morphology. To address both noise and background variability in robot self-modeling, we propose a task-aware denoising pipeline (Fig.~\ref{Pipeline}) that operates in three steps: (i) semantic segmentation to isolate the robot, (ii) integration of the clean binary mask into the self-modeling process, and (iii) noise-specific filtering—Wiener filtering, median filtering, and NLM combined with IFT-SVM classification. This ordering ensures that the robot is first separated from the background and correctly modeled, before denoising restores corrupted inputs while preserving morphological fidelity. As a result, morphological features remain detectable and reconstructible even under high-noise and visually cluttered environments conditions to be subsequently transferred to a self-modeling engine based on FFKSM. 




\subsection{Semantic Segmentation for Complex Backgrounds}

Current self-modeling approaches, such as FFKSM, rely on color-based segmentation against plain white backgrounds, an assumption that fails in realistic, cluttered, and colorful environments. To address this, we integrate semantic segmentation to isolate the robot from cluttered scenes. For each image $I \in \mathbb{R}^{H \times W \times 3}$, we define a mask $M \in \{0,1\}^{H \times W}$ where 1 indicates the robot and 0 the background.

\subsubsection{\textbf{Model Architecture}}
We employ a deep convolutional network (FBV\_SM) optionally augmented with a positional encoder \cite{32,33} to capture fine spatial details:

\begin{equation*}
\mathbf{z} = \gamma(\mathbf{x})=
\end{equation*}
\begin{equation}
 [\mathbf{x}, \sin(2^0 \pi \mathbf{x}), \cos(2^0 \pi \mathbf{x}), \dots, \sin(2^{L-1} \pi \mathbf{x}), \cos(2^{L-1} \pi \mathbf{x})],
\end{equation}
where $\mathbf{x}=(x,y)$ are normalized coordinates, and $L$ is the number of frequency bands. These embeddings are concatenated with RGB features and passed to FBV\_SM, producing per-pixel logits:

\begin{equation}
\mathbf{y}_{i,j} = f_\theta(I_{i,j}, \mathbf{z}_{i,j}) \in \mathbb{R}^{C},
\end{equation}
where $C = 2$ denotes the number of classes (background and robot), and $\theta$ represents the network parameters.

\subsubsection{\textbf{Training Objective}}
The network is trained using the cross-entropy loss:

\begin{equation}
\mathcal{L}(\theta) = - \frac{1}{H W} \sum_{i=1}^{H} \sum_{j=1}^{W} \sum_{c=0}^{1} \mathbf{1}_{[M_{i,j}=c]} \log \frac{\exp(\mathbf{y}_{i,j,c})}{\sum_{k=0}^{1} \exp(\mathbf{y}_{i,j,k})},
\end{equation}
which encourages accurate per-pixel classification. Optimization is performed using Adam with a learning rate selected empirically.

\subsubsection{\textbf{Inference and Mask Generation}}
During inference, the predicted mask $\hat{M}$ is obtained by selecting the class with the highest logit at each pixel:

\begin{equation}
\hat{M}_{i,j} = \arg\max_c \mathbf{y}_{i,j,c}.
\end{equation}

The mask can be optionally colorized (e.g., blue for background, red for robot) for visualization purposes, providing qualitative insights into segmentation performance.

\subsubsection{\textbf{Integration with Denoising and Self-Modeling}}
The predicted mask is applied to the denoised image $I_\text{denoise}$ obtained from the techniques in Section~\ref{sec:denoising_techniques}:

\begin{equation}
I_\text{seg} = I_\text{denoise} \odot \hat{M},
\end{equation}
where $\odot$ denotes element-wise multiplication. This operation removes background clutter while preserving the robot’s morphological and kinematic features, providing high-fidelity input for the self-modeling module.

\subsection{Wiener Filter for Blur Removal}

Blurring reduces edge sharpness and obscures joint boundaries critical for kinematic reconstruction. The Wiener filter \cite{23,24} reverses blur in the frequency domain while suppressing noise amplification:

\begin{equation}
\hat{I}(u,v) = \frac{H^*(u,v)}{|H(u,v)|^2 + S_n(u,v)/S_i(u,v)} I_\text{blur}(u,v),
\end{equation}
where $H(u,v)$ is the Fourier transform of the blur kernel, and $S_n(u,v)$ and $S_i(u,v)$ denote noise and image power spectra.  
By estimating the degradation function, the Wiener filter sharpens edges and restores fine details while suppressing the amplification of residual noise, making it particularly effective for motion or defocus blur.

\subsection{Median Filter for Salt-and-Pepper Noise}

Impulsive disturbances from salt-and-pepper noise disrupt visual cues. Median filtering \cite{25,26} removes such noise while preserving edges:

\begin{equation}
I_\text{denoise}(x,y) = \text{median}\Big\{ I_\text{sp}(i,j) \,|\, (i,j) \in \mathcal{N}_{k \times k}(x,y) \Big\},
\end{equation}
where $\mathcal{N}_{k \times k}$ is a local neighborhood.  
Because salt-and-pepper noise appears as isolated extreme pixel values, the median filter replaces outliers with representative local intensities, effectively removing noise while keeping structural boundaries intact.


\subsection{Non-Local Means Denoising with Intuitionistic Fuzzy Twin Support Vector Machines}

Gaussian noise in images is commonly addressed using Non-Local Means (NLM) \cite{27,28}, which denoises each pixel by averaging similar patches across the image:

\begin{equation}
I_\text{denoise}(x) = \sum_{y \in \Omega} w(x, y) I_\text{art}(y),
\end{equation}
with weights:
\begin{equation*}
w(x,y) = \frac{1}{Z(x)} \exp\Big( - \frac{|I(\mathcal{N}_x) - I(\mathcal{N}_y)|^2_2}{h^2} \Big),
\end{equation*}

\begin{equation}
Z(x) = \sum_{y \in \Omega} \exp(\cdot),
\end{equation}
where $\mathcal{N}_x$ and $\mathcal{N}_y$ denote image patches centered at $x$ and $y$, and $h$ is a smoothing parameter. NLM effectively preserves edges and fine textures while removing Gaussian noise. Since Gaussian noise affects all pixels with small random variations, NLM leverages self-similarity across the image to suppress noise while keeping fine structural details.

While standard NLM provides strong denoising, its performance can degrade under high noise or in regions with complex textures. To enhance robustness, we integrate Intuitionistic Fuzzy Twin Support Vector Machines (IFTSVMs) \cite{36}, which assign both membership and non-membership degrees to image pixels, modeling uncertainty caused by noise. After NLM filtering, IFTSVM adaptively refines each patch by reducing residual noise while preserving edges and textures, leading to improved overall image quality.

\subsection{Self-modeling Engine}

In this work, we adopt the FFKSM framework \cite{1}, which reconstructs a robot’s body schema by aligning observed motion trajectories with candidate kinematic models. Specifically, FFKSM leverages clean binary masks obtained under controlled white backgrounds of the robot across different poses, infers joint positions, and iteratively searches for the kinematic configuration that best explains the observed motion. FFKSM formulates self-modeling using three neural networks: a coordinate network that encodes spatial positions of the robot body, a kinematic network that maps joint configurations to morphological structure, and a prediction network that estimates the distribution parameters density and distance for reconstructing the robot’s shape. Together, these components enable the robot to iteratively refine its internal self-model from ideal visual observations.

\section{Experimental Setup}\label{exp-setup}

To rigorously evaluate the robustness of self-modeling under realistic visual conditions, we conducted experiments along two complementary axes: (1) replicating the original FFKSM experiments using their dataset and code, and (2) validating transferability on a 3D-printed robot with new datasets collected under realistic, cluttered, and noisy conditions. In both cases, we systematically introduced noise and applied our proposed denoising and segmentation pipeline to assess its ability to overcome the limitations of the baseline visual self-modeling approach. All experiments were executed on a workstation equipped with an Intel Core i9 CPU and 32 GB RAM running Ubuntu Linux, ensuring reproducibility and stable runtime performance.


\subsection{Replication of Original Experiments}

We first reproduced the experiments of Hu \textit{et al.} \cite{1} using their publicly available code and dataset. This established a baseline performance of the FFKSM pipeline in its intended, noise-free, white-background setting.

However, this baseline setting highlights a key limitation: the original pipeline assumes ideal imaging conditions and avoids depth sensing due to noise sensitivity. To expose this weakness, we injected three types of Gaussian blur, salt-and-pepper, and artificial/Gaussian noise—into the original dataset and re-ran the FFKSM pipeline. These perturbations significantly degraded morphology reconstruction and trajectory prediction, showing that the original approach is brittle under non-ideal conditions.

\subsection{Experiments on 3D-Printed Robot}

\begin{figure}[htbp]
    \centering
    \begin{subfigure}[b]{0.265\textwidth} 
        \includegraphics[width=\textwidth]{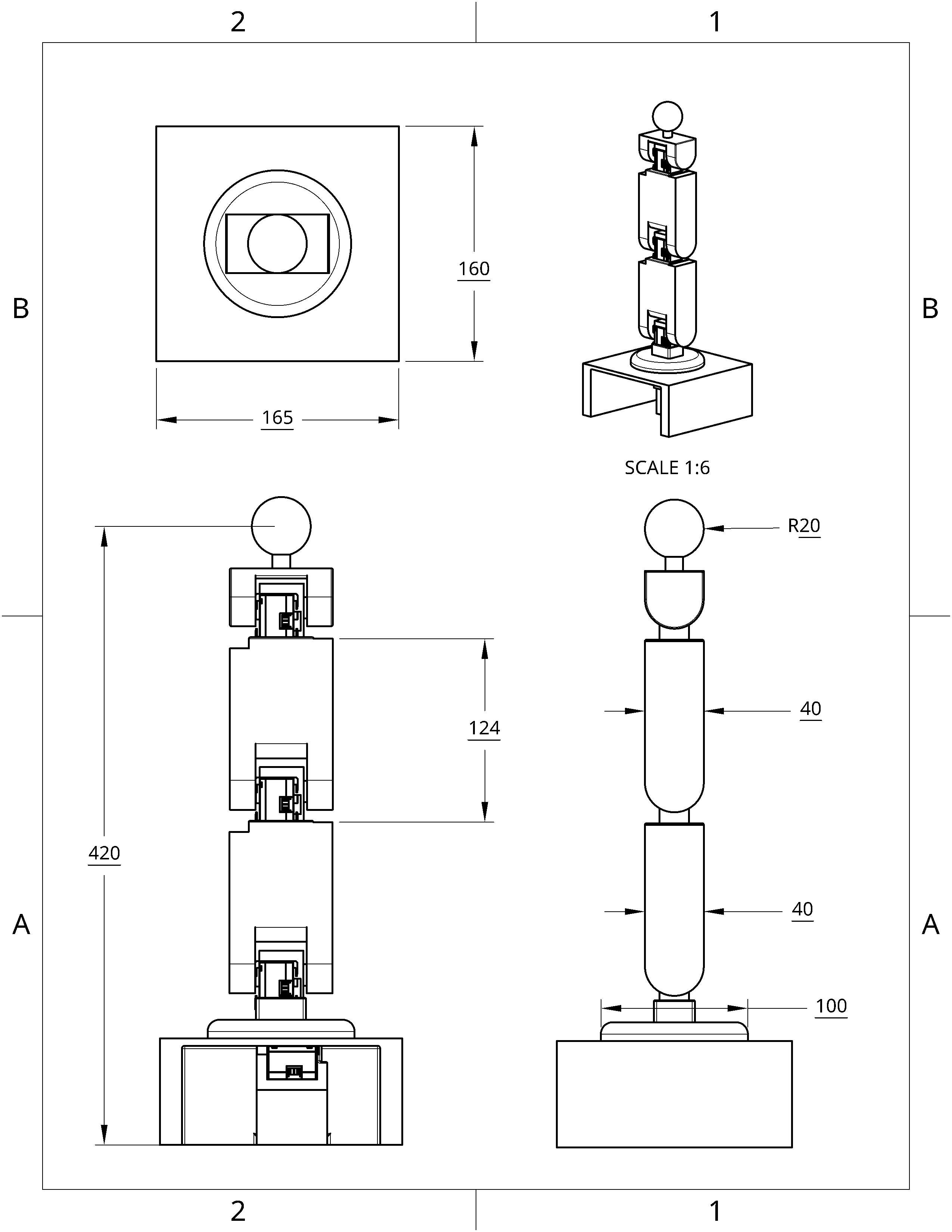} %
        \caption{Mechanical design}
        \label{fig:figure1}
    \end{subfigure}
    \hfill 
    \begin{subfigure}[b]{0.205\textwidth} 
        \includegraphics[width=\textwidth]{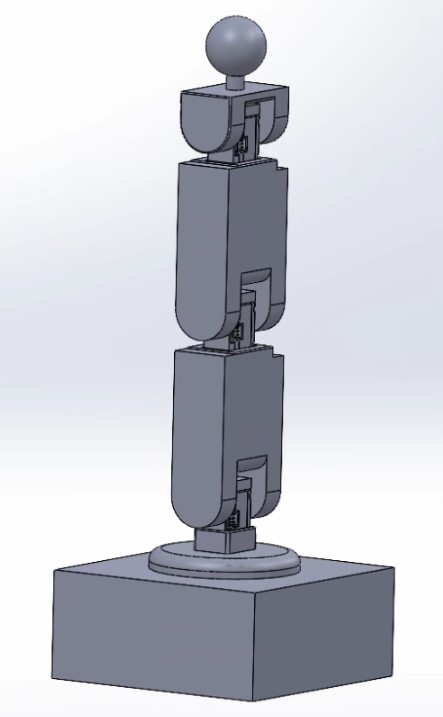} %
        \caption{Robot}
        \label{fig:figure2}
    \end{subfigure}
    \caption{Mechanical design of the printed robot highlighting its structural layout}
    \label{fig:overall_figure}
\end{figure}

To move beyond the constrained digital dataset of~\cite{1}, we fabricated a 4-degree-of-freedom (DOF) robotic manipulator using 3D-printed PLA components and Dynamixel XL330-M288 servos (Figure~\ref{fig:overall_figure}). The platform preserved the kinematic structure of the original system while introducing realistic sources of variation, including mechanical tolerances, material imperfections, and actuator variability, thereby providing a more challenging testbed for self-modeling than the strictly synthetic setting of the FFKSM study. The kinematic chain consisted of a rotating base, two intermediate links, and a terminal end-effector, each with $\pm 90^\circ$ rotation.

\begin{figure}[htbp]
    \centering
    \begin{subfigure}[b]{0.189\textwidth} 
        \includegraphics[width=\textwidth]{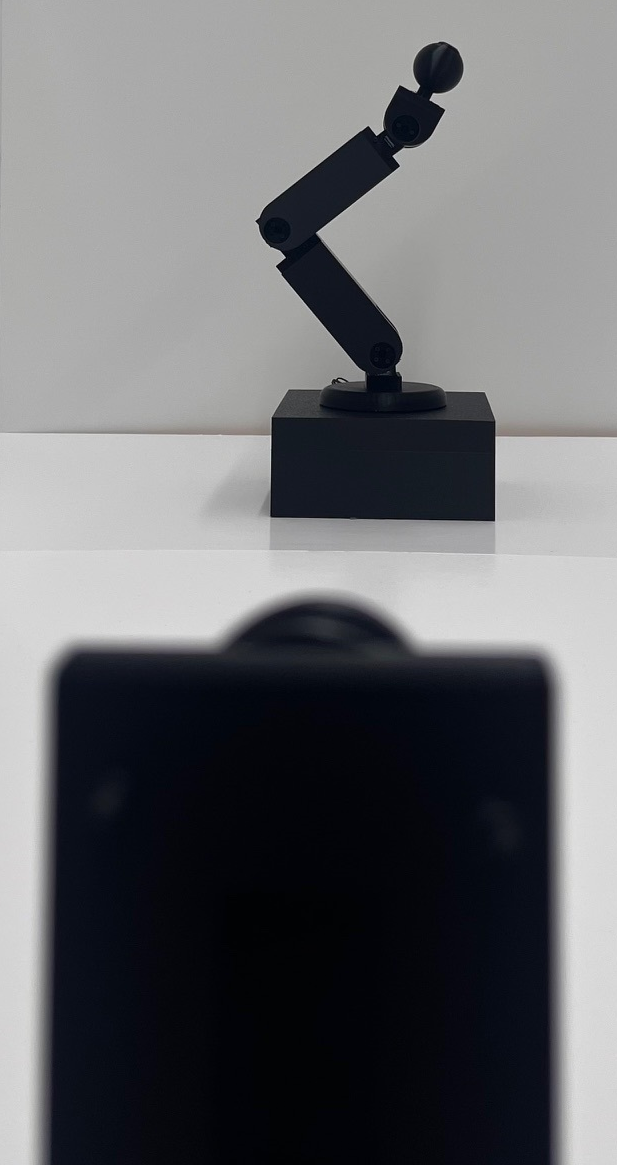} %
        \caption{Robot and camera}
        \label{fig:figure1}
    \end{subfigure}
    \hfill 
    \begin{subfigure}[b]{0.25\textwidth} 
        \includegraphics[width=\textwidth]{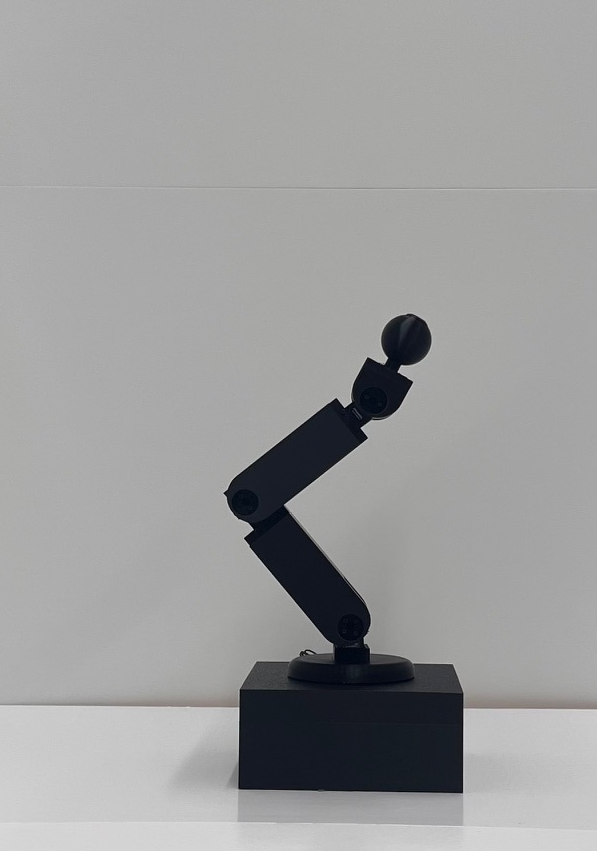} %
        \caption{Robot with background}
        \label{fig:figure2}
    \end{subfigure}
    \caption{Experimental imaging setup with the robot positioned in front of the camera}
    \label{fig:overall_figure1}
\end{figure}

For vision-based evaluation, the base was printed in white PLA to be excluded from segmentation, while all movable links were printed in black PLA to maximize contrast. RGB images were captured at 640×480 resolution using an Intel RealSense D435 camera and then downsampled to 100×100 pixels for consistency with Hu \textit{et al.}~\cite{1} (Figure~\ref{fig:overall_figure1}). From the captured dataset of 12,000 images, 10,000 were used for training and 2,000 for testing.

Unlike the original study, which assumed a perfectly white background, we additionally evaluated the robot under cluttered and colorful scenes. In these settings, the FFKSM color-segmentation approach failed, while our semantic segmentation pipeline reliably isolated the robot, demonstrating robustness to realistic visual conditions. Finally, small variations in servo precision and printed geometry introduced mild but consistent noise into the dataset, further aligning the test scenario with real-world deployment challenges.

\subsection{Noise Injection Scenarios}

For both datasets (original and 3D-printed robot), we systematically injected the following types of visual noise:

\begin{enumerate}
\item Gaussian blur – images were convolved with a Gaussian kernel of varying standard deviation.
\item Salt-and-pepper noise – random pixels were replaced with black or white values at varying densities.
\item Zero-mean Gaussian noise – added to pixel intensities with varying variance.
\end{enumerate}

These scenarios allow us to quantify how noise propagates through the self-modeling pipeline and assess the effectiveness of denoising.

\subsection{Denoising Pipeline Evaluation}
We applied the proposed task-aware denoising framework (Wiener filtering, median filtering, and Non-Local Means) to all noisy datasets. Semantic segmentation was also applied to relax the white-background assumption, enabling evaluation in cluttered or colorful environments. Denoised images were fed into the FFKSM pipeline for morphology and dynamics reconstruction.

\subsection{Evaluation Metrics}

To rigorously assess self-modeling performance under different noise and background conditions, we employ both quantitative and qualitative metrics:

\subsubsection{Morphology Reconstruction Error}
We quantify the difference between predicted and ground-truth 3D structure using mean squared error (MSE):

\begin{equation}
\mathrm{MSE} = \frac{1}{N} \sum_{i=1}^{N} \|\mathbf{X}_i^\text{pred} - \mathbf{X}_i^\text{gt}\|^2,
\end{equation}

where $N$ is the number of 3D points, $\mathbf{X}_i^\text{pred}$ is the predicted 3D coordinate of point $i$, and $\mathbf{X}_i^\text{gt}$ is the corresponding ground-truth coordinate.
Lower MSE indicates more accurate morphology reconstruction. We report MSE for the noise-free baseline and for each noise condition.

\subsubsection{Semantic Segmentation Quality}
To isolate the robot from cluttered or colorful backgrounds, we compute the Intersection over Union (IoU) between the predicted mask $\hat{M}$ and the ground-truth mask $M$:
\begin{equation}
\mathrm{IoU} = \frac{|\hat{M} \cap M|}{|\hat{M} \cup M|}.
\end{equation}
IoU values close to 1 indicate accurate segmentation. While the original FFKSM pipeline assumes a white background and fails in cluttered scenes, we provide qualitative comparisons to demonstrate the effectiveness of our semantic segmentation module in realistic environments.

\subsubsection{F1-Score}
To complement IoU, we also report the F1-score \cite{36}, which balances precision and recall into a single measure of segmentation quality. Precision evaluates the fraction of correctly predicted foreground pixels out of all predicted foreground pixels, while recall measures the fraction of correctly predicted foreground pixels out of all actual foreground pixels. The F1-score is defined as the harmonic mean of precision and recall:

\begin{equation}
\mathrm{Precision} = \frac{TP}{TP + FP}, \quad
\mathrm{Recall} = \frac{TP}{TP + FN},
\end{equation}

\begin{equation}
\mathrm{F1} = \frac{2 \cdot \mathrm{Precision} \cdot \mathrm{Recall}}{\mathrm{Precision} + \mathrm{Recall}},
\end{equation}

where $TP$, $FP$, and $FN$ denote true positives, false positives, and false negatives, respectively.
F1-scores close to 1 indicate that both precision and recall are high, meaning that the segmentation correctly captures the robot body while avoiding false detections.

\subsubsection{Comparison Across Conditions}
Metrics are reported for the original dataset, 3D-printed robot, noisy vs denoised inputs, and white vs cluttered backgrounds. This allows systematic evaluation of the impact of noise, denoising, and segmentation on morphology reconstruction, trajectory prediction, and downstream task performance.

\section{Experimental Results}\label{exp-results}

We evaluate our self-modeling pipeline on challenging visual scenarios, comparing it with FFKSM. Experiments include semantic segmentation in cluttered backgrounds, baseline performance on noise-free data, and robustness under various noise types. Results demonstrate improved accuracy, noise resilience, and reliable generalization to real-world conditions.

\subsection{Semantic Segmentation in Cluttered Backgrounds}

\begin{figure}[!t]
\centering
\includegraphics[width=3.3in, height=1in]{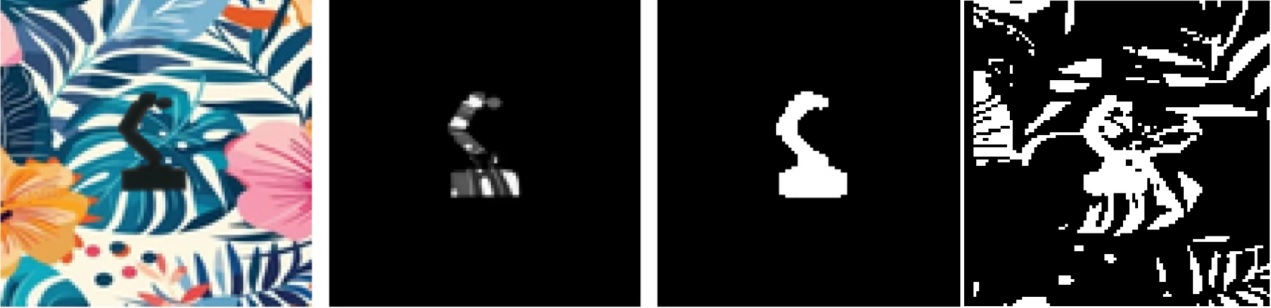}
\caption{Semantic segmentation comparison on a colorful leaf background.}
\label{Seg}
\end{figure}

\begin{figure}[!t]
\centering
\includegraphics[width=3.3in, height=1in]{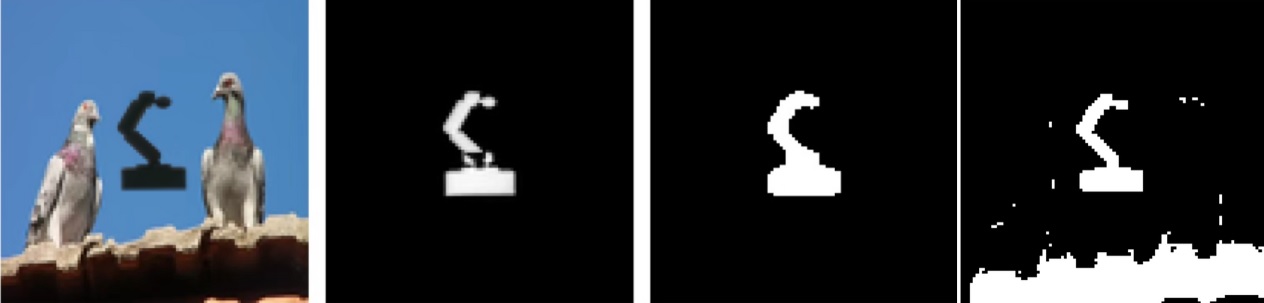}
\caption{Semantic segmentation comparison with two gray pigeons in the background.}
\label{Seg2}
\end{figure}

\begin{figure}[!t]
\centering
\includegraphics[width=3.3in, height=1in]{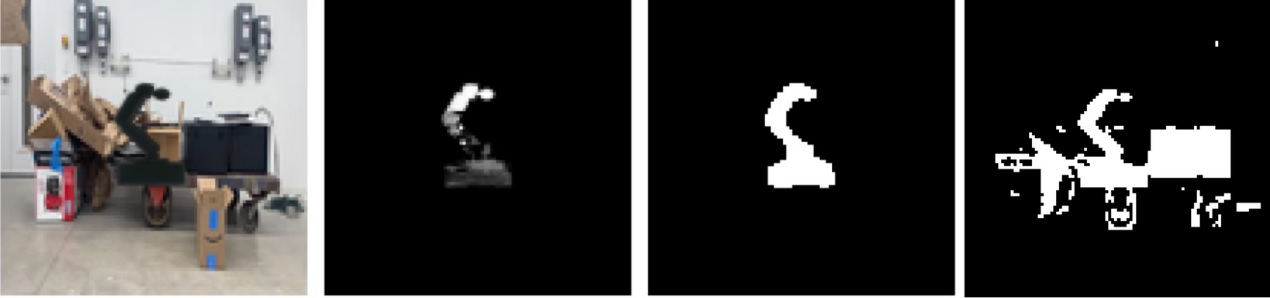}
\caption{Semantic segmentation comparison in the lab environment.}
\label{Seg3}
\end{figure}

Traditional color-based segmentation, as used in FFKSM, is unreliable in complex, cluttered environments. By leveraging motion-based and semantic segmentation, our pipeline consistently identifies the robot across diverse backgrounds, achieving over fourfold improvement in both IoU and F1-score compared to FFKSM. This ensures robust downstream morphology and dynamics reconstruction under realistic conditions.

i) \textbf{Colorful leaf background:} FFKSM failed due to color-based binary segmentation being confused by the leaves, rendering the robot nearly indistinguishable (IoU: 0.1645, F1: 0.2826). Our semantic segmentation, aided by motion cues, produced a clear binary mask with significantly higher accuracy (IoU: 0.7070, F1: 0.8283), making the robot fully visible (See Fig.~\ref{Seg}, Table.~\ref{tab:seg_results}).

ii) \textbf{Colorful leaf background:} FFKSM failed due to color-based binary segmentation being confused by the leaves, rendering the robot nearly indistinguishable (IoU: 0.1645, F1: 0.2826). Our semantic segmentation, aided by motion cues, produced a clear binary mask with significantly higher accuracy (IoU: 0.7070, F1: 0.8283), making the robot fully visible (See Fig.~\ref{Seg}, Table.~\ref{tab:seg_results}).

iii) \textbf{Two gray pigeons background:} FFKSM partially recognized the robot but incorrectly included roof pixels as part of the mask (IoU: 0.1518, F1: 0.2636). In contrast, our method correctly segmented the robot from the background, yielding a precise and complete mask (IoU: 0.6690, F1: 0.8017) (See Fig.~\ref{Seg2}, Table.~\ref{tab:seg_results}).

iv) \textbf{Lab environment background:} FFKSM was confused by dark objects overlapping the robot, producing incorrect segmentation (IoU: 0.2531, F1: 0.4040). Our method reliably separated the robot from all background clutter, producing a fully accurate mask (IoU: 0.6729, F1: 0.8045) (See Fig.~\ref{Seg3}, Table.~\ref{tab:seg_results}).

These results demonstrate that traditional color-based segmentation, as used in FFKSM, is unreliable in complex, cluttered environments. By leveraging motion-based and semantic segmentation, our pipeline consistently identifies the robot across diverse backgrounds, achieving up to \textbf{4$\times$ improvement in IoU} and \textbf{3$\times$ improvement in F1-score}, enabling robust downstream morphology and dynamics reconstruction.

\begin{figure}[!t]
\centering
\includegraphics[width=3.3in, height=1in]{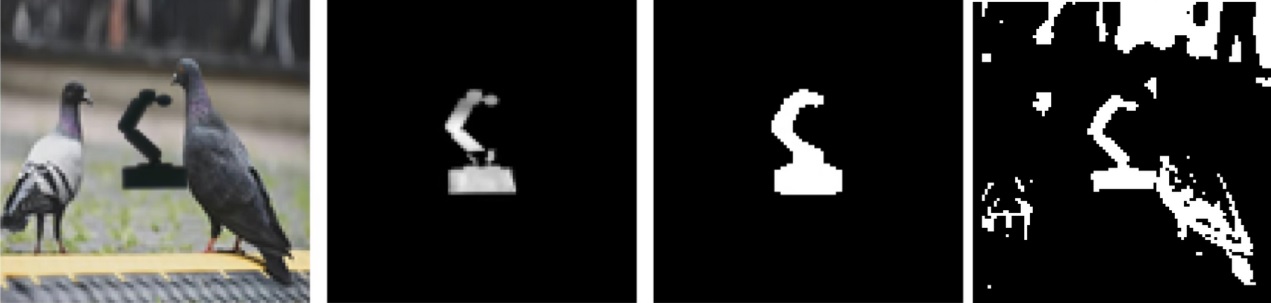}
\caption{Semantic segmentation comparison with two black pigeons in the background.}
\label{Seg1}
\end{figure}

\begin{table}[!t]
\centering
\caption{Comparison of segmentation performance (IoU and F1-score) between FFKSM and our method across different backgrounds.}
\label{tab:seg_results}
\begin{tabular}{lcccc}
\hline
\multirow{2}{*}{\textbf{Background}} & \multicolumn{2}{c}{\textbf{FFKSM}} & \multicolumn{2}{c}{\textbf{Our Method}} \\
\cline{2-5}
& IoU & F1-score & IoU & F1-score \\
\hline
Leaf Background   & 0.1645 & 0.2826 & 0.7070 & 0.8283 \\
Gray Pigeons      & 0.1518 & 0.2636 & 0.6690 & 0.8017 \\
Black Pigeons     & 0.1556 & 0.2693 & 0.7027 & 0.8254 \\
Lab Background    & 0.2531 & 0.4040 & 0.6729 & 0.8045 \\
\hline
\end{tabular}
\end{table}

\subsection{Baseline Performance}

\begin{figure}[!t]
\centering
\includegraphics[width=2.8in, height=1.9in]{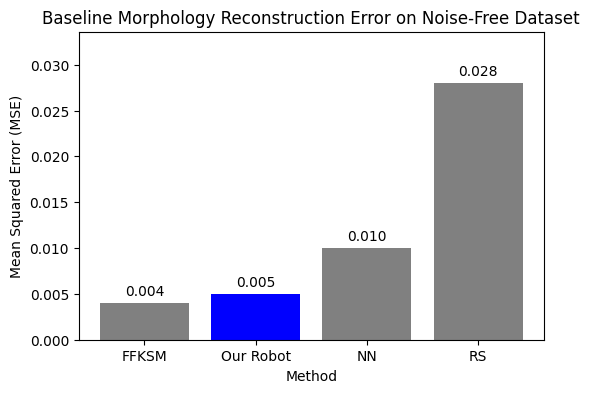}
\caption{Baseline morphology reconstruction error on the noise-free dataset. Our robot (blue) is compared with FFKSM, Nearest Neighbor, and Random Selection.}
\label{Robot5}
\end{figure}

On the noise-free original dataset, the FFKSM pipeline reproduced the morphology and trajectory results reported in \cite{1}, validating our reimplementation. The 3D-printed robot exhibited slightly higher reconstruction error due to geometric and calibration variations, but remained close to baseline. These results provide a reference point for subsequent noise and denoising experiments (Fig.~\ref{Robot5}).

\subsection{Comparison Across Datasets}

By default, FFKSM cannot handle noisy inputs and exhibits sharp performance degradation under all three corruptions. For a fair comparison, we manually applied task-aware denoising to FFKSM, which allowed it to partially recover and approach its noise-free baseline. In contrast, our pipeline intrinsically integrates noise-robust segmentation and denoising, handling blur, salt-and-pepper, and Gaussian noise without additional intervention and consistently achieving near-baseline performance.

We next evaluated robustness under three representative noise types: Gaussian blur, salt-and-pepper, and Gaussian (artificial) noise. Across both the synthetic and 3D-printed datasets, noise alone substantially degraded performance for FFKSM and our method, with MSE values rising three- to fivefold compared to the noise-free baseline. Blurring softened joint boundaries, salt-and-pepper introduced impulsive artifacts, and Gaussian noise disrupted pixel intensities, all reducing morphology fidelity.

Applying our task-aware denoising pipeline effectively restored structure in each case: Wiener filtering for blur, median filtering for salt-and-pepper, and Non-Local Means with IFT-SVM for Gaussian noise. Post-denoising, both pipelines recovered near-baseline accuracy, with our method consistently achieving slightly lower error (0.0054–0.0059) compared to FFKSM (0.0061–0.0064). These results confirm that noise severely impairs visual self-modeling, but task-specific denoising reliably mitigates its impact.

\begin{figure}[!t]
\centering
\includegraphics[width=3.3in, height=2in]{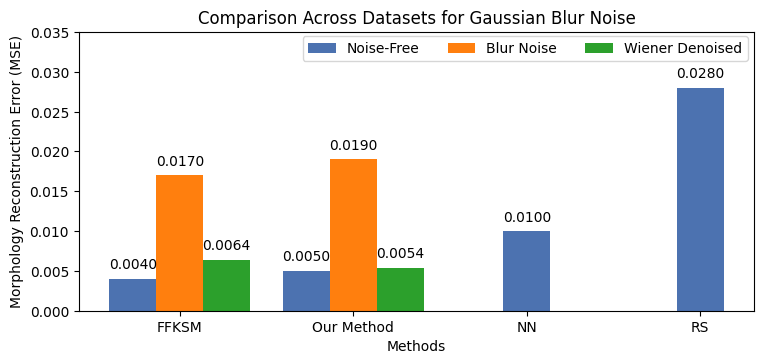}
\caption{Comparison of morphology reconstruction error (MSE) under Gaussian blur with and without Wiener filtering across datasets.}
\label{Comp3}
\end{figure}

\begin{figure}[!t]
\centering
\includegraphics[width=3.3in, height=2in]{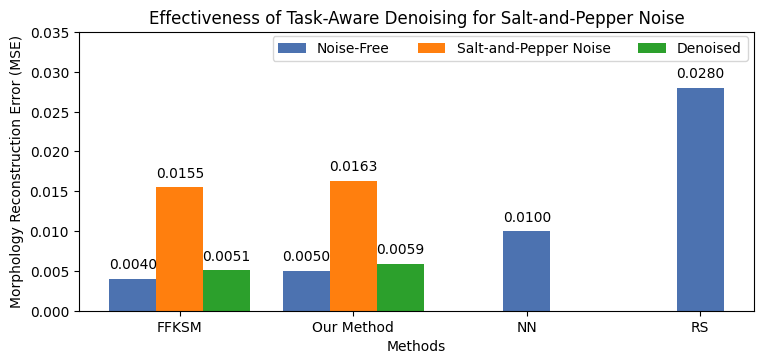}
\caption{Comparison of morphology reconstruction error (MSE) under salt-and-pepper noise with and without median filtering across datasets.}
\label{Comp}
\end{figure}

\begin{figure}[!t]
\centering
\includegraphics[width=3.3in, height=2in]{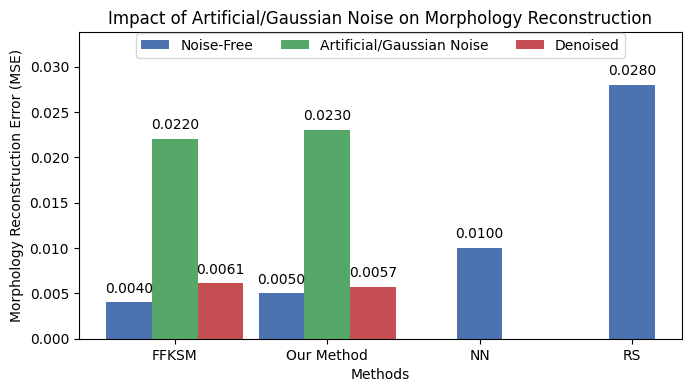}
\caption{Comparison of morphology reconstruction error (MSE) under Gaussian noise with and without Non-Local Means denoising across datasets.}
\label{Comp1}
\end{figure}

Figures~\ref{Comp3}–\ref{Comp1} summarize these trends, illustrating baseline, noisy, and denoised reconstruction error for each corruption type. Performance was consistent across the original synthetic dataset and the 3D-printed robot, demonstrating generalization to realistic conditions with mechanical and visual variability. For context, naive baselines such as nearest-neighbor (MSE 0.010) and random selection (MSE 0.028) remained substantially worse than either self-modeling pipeline.

Overall, these results highlight three key points: (i) visual noise significantly impairs morphology reconstruction, (ii) task-aware denoising restores near-baseline performance, and (iii) our method generalizes across datasets and outperforms naive strategies, providing a robust path for self-modeling in real-world deployments.

\section{Conclusion}\label{concl}

We have presented the first systematic study of visual noise effects on robotic self-modeling, encompassing blur, salt-and-pepper, and artificial/Gaussian noise. Our task-aware denoising pipeline—combining Wiener filtering, median filtering, Non-Local Means, and advanced integration with IFT-SVM—effectively restores morphology accuracy under these adverse conditions, demonstrating that visual noise alone can severely degrade self-modeling if unaddressed. Quantitative evaluations show substantial improvements in robot morphology reconstruction across both synthetic and real-world datasets. Furthermore, by integrating semantic segmentation, our method reliably handles cluttered and colorful real-world environments, overcoming the limitations of prior pipelines that assume a uniform background. Together, these contributions highlight that both noise mitigation and background robustness are essential for high-fidelity, deployable self-modeling in robotics, enabling more reliable perception and autonomous adaptation in complex environments.


\bibliographystyle{IEEEtran}
\bibliography{mybib}

\end{document}